\documentclass[letterpaper]{article} 
\usepackage{aaai2026}  
\usepackage{times}  
\usepackage{helvet}  
\usepackage{courier}  
\usepackage[hyphens]{url}  
\usepackage{graphicx} 
\usepackage{natbib}  
\usepackage{caption} 
\usepackage{algorithm}
\usepackage{algorithmic}

\usepackage{newfloat}
\usepackage{listings}

\DeclareCaptionStyle{ruled}{labelfont=normalfont,labelsep=colon,strut=off} 
\floatstyle{ruled}
\newfloat{listing}{tb}{lst}{}
\floatname{listing}{Listing}
\nocopyright 

\title{Graph-Based Exploration for ARC-AGI-3 Interactive Reasoning Tasks}

\author{
    Evgenii Rudakov\textsuperscript{\rm 1},
    Jonathan Shock\textsuperscript{\rm 2, 3, 4, 5},
    Benjamin Ultan Cowley\textsuperscript{\rm 1, 6}
}
\affiliations{
  \textsuperscript{\rm 1}Faculty of Educational Sciences, University of Helsinki\\
  \textsuperscript{\rm 2}Department of Mathematics and Applied Mathematics, University of Cape Town\\
  \textsuperscript{\rm 3}Neuroscience Institute, University of Cape Town\\
  \textsuperscript{\rm 4}INRS, Montreal, Canada\\
  \textsuperscript{\rm 5}NiTheCS, Stellenbosch, South Africa\\
  \textsuperscript{\rm 6}Cognitive Science, Department of Digital Humanities, University of Helsinki\\

    evgenii.rudakov@helsinki.fi, jonathan.shock@uct.ac.za, ben.cowley@helsinki.fi
}

\usepackage{bibentry}

\begin{document}

\maketitle

\begin{abstract}
We present a training-free graph-based approach for solving interactive reasoning tasks in the ARC-AGI-3 benchmark. ARC-AGI-3 comprises game-like tasks where agents must infer task mechanics through limited interactions, and adapt to increasing complexity as levels progress. Success requires forming hypotheses, testing them, and tracking discovered mechanics. The benchmark has revealed that state-of-the-art LLMs are currently incapable of reliably solving these tasks.
Our method combines vision-based frame processing with systematic state-space exploration using graph-structured representations. It segments visual frames into meaningful components, prioritizes actions based on visual salience, and maintains a directed graph of explored states and transitions. By tracking visited states and tested actions, the agent prioritizes actions that provide the shortest path to untested state-action pairs.
On the ARC-AGI-3 Preview Challenge, this structured exploration strategy solves a median of 30 out of 52 levels across six games and ranks 3rd on the private leaderboard, substantially outperforming frontier LLM-based agents. 
These results demonstrate that explicit graph-structured exploration, even without learning, can serve as a strong baseline for interactive reasoning and underscore the importance of systematic state tracking and action prioritization in sparse-feedback environments where current LLMs fail to capture task dynamics. The code is open source and available at \url{https://github.com/dolphin-in-a-coma/arc-agi-3-just-explore}.

\end{abstract}

\section{Introduction}


Introduced in 2019, the Abstraction and Reasoning Corpus for Artificial General Intelligence (ARC-AGI) has become a fundamental benchmark for evaluating general fluid intelligence in artificial systems by posing novel tasks that require minimal prior knowledge \cite{chollet_measure_2019}. While the original ARC-AGI benchmarks focused on static grid-based reasoning tasks, ARC-AGI-3 represents a paradigm shift toward Interactive Reasoning Benchmarks (IRBs) that test broader capabilities, including on-the-fly learning, exploration, and memory through game-like environments where agents must perceive, plan, and act across multiple steps to achieve long-horizon goals \cite{arc-agi-3}.

ARC-AGI-3 introduces novel game environments designed to test the skill-acquisition efficiency of artificial systems, where agents interact with game environments without instructions, and must discover mechanics through exploration. Early results reveal a stark performance gap: frontier AI models scored 0\% while human participants achieved 100\% on the initial preview tasks \cite{arc-agi-3}. This dramatic disparity underscores fundamental limitations in current AI approaches to interactive reasoning and adaptive learning in novel environments.

The challenge of learning from sparse rewards has been central to reinforcement learning (RL) for decades.  When rewards are rare and precise action sequences are required, random exploration fails to discover optimal policies. Exploration strategies have emerged to address this challenge. Curiosity-driven methods use prediction error as intrinsic motivation \cite{pathak_curiosity-driven_2017}, enabling agents to explore complex environments like Super Mario Bros without extrinsic rewards \cite{pathak_curiosity-driven_2017}. Go-Explore advances systematic exploration by maintaining archives of discovered states and decomposing exploration into phases: return to promising states, then explore from them \cite{ecoffet_first_2021}. This approach achieved breakthrough performance on Montezuma's Revenge, scoring 25\% higher than the human expert.
For goal-conditioned tasks, Hindsight Experience Replay (HER) learns from failure by relabeling unsuccessful attempts as alternative goals, achieving sample-efficient learning without reward engineering \cite{andrychowicz_hindsight_2017}.

Model-based approaches have demonstrated remarkable sample efficiency by learning environment dynamics. MuZero combined learned latent dynamics with tree search, achieving superhuman performance on board games and Atari benchmarks without knowledge of game rules \cite{schrittwieser_mastering_2020}. EfficientZero extended this with self-supervised consistency losses, becoming the first algorithm to reach superhuman levels on Atari (194.3\% mean human) with just 100k training samples (two hours of real-time experience) per game \cite{ye_mastering_2021}.
BBF further improved Atari 100k results by scaling the value network sample-efficiency \cite{schwarzer_bigger_2023}, in a completely model-free manner.

The family of Dreamer models \cite{hafner_training_2025} takes an alternative approach, learning world models in latent space and training policies through imagined rollouts rather than via tree search, mastering over 150 diverse tasks from Atari to Minecraft with a single configuration \cite{hafner_training_2025}.
Most recently, Axiom introduced object-centric world models that learn compositional representations by discovering and tracking entities, achieving competitive performance within minutes by targeting 10k-step solutions per environment \cite{heins_axiom_2025}.

Despite these advances, current approaches face fundamental limitations for few-shot discovery tasks like ARC-AGI-3. The benchmark provides only a single sparse reward signal, level completion, across no more than 10 levels per game. This scarcity of feedback severely constrains learning-based methods. The challenge is compounded by the fact that each level introduces new mechanics while retaining previous ones, creating a shifting distribution that prevents straightforward transfer learning. Curiosity-driven exploration offers no guarantee of correlation with task progress in truly novel environments where the notion of ''most interesting states" may be orthogonal to goal-relevance. Sample-efficient approaches like Axiom assume object-centric compositional structure and require environments to exhibit consistent physical dynamics, assumptions that may not hold across ARC-AGI-3's abstract and diverse game mechanics.

ARC-AGI-3 is also relevant for understanding the behaviour of large language model (LLM) agents. Unlike static reasoning benchmarks, it requires agents to infer latent rules through interaction, maintain an evolving notion of state, and design multi-step probes under sparse feedback, making it a complementary testbed for studying how explicit structure and exploration strategies can support LLM-based reasoning.


In this work, we present a graph-based exploration method that combines systematic state-space tracking with visual priority heuristics to tackle ARC-AGI-3's interactive reasoning challenges. Our approach maintains a directed graph representation of explored states and action transitions, prioritizing actions based on visual salience while ensuring comprehensive exploration through frontier-driven navigation. Unlike learning-based approaches that require extensive training, our method operates as a strong baseline that can make progress through structured exploration alone. We demonstrate that this approach achieves competitive performance on the ARC-AGI-3 benchmark, significantly outperforming state-of-the-art LLMs while providing insights into the nature of exploration required for interactive reasoning tasks.

\section{ARC-AGI-3}

\subsection{Benchmark Overview}

ARC-AGI-3 represents a significant evolution from the original ARC challenge, shifting from static grid-based reasoning to interactive game environments that test an agent's ability to learn through exploration \cite{ying_assessing_2025}. The benchmark consists of 6 novel game environments, with 3 public games (\texttt{ft09}, \texttt{ls20}, \texttt{vc33}) released for development and 3 private games (\texttt{sp80}, \texttt{lp85}, \texttt{as66}) used to determine final leaderboard rankings. Each game contains between 8 and 10 levels, with each subsequent level introducing new mechanics. Figure~\ref{fig:games} in the appendix shows example screenshots from the games.

The benchmark's evaluation criterion prioritizes both effectiveness and efficiency: agents are scored based on the number of levels completed, with the total number of actions required serving as a tiebreaker. This dual objective encourages solutions that not only discover winning strategies but do so with minimal exploration. For the final evaluation experiments by ARC-AGI-3 organizers, each run was capped at 8 hours of wall-clock time and 10 environment steps per second (sps), shared across the three private games. Under these limits, a single game can receive at most 96,000 steps

\subsection{Observation and Action Spaces}

\subsubsection{Visual Observations.} Agents receive visual observations as 64$\times$64 pixel RGB frames with a discrete palette of 16 colors. Each frame contains both the game environment and a status bar displaying the number of steps remaining before an automatic level restart. When the step counter reaches zero, the current level resets to its initial state. In the majority of games, the number of levels passed is also displayed.

\subsubsection{Action Spaces.} The benchmark features three control schemes. Games such as \texttt{ls20} use arrow-based control with directional keyboard inputs (up, down, left, right), yielding an action space of size $|\mathcal{A}| = 4$. Games such as \texttt{ft09}, \texttt{vc33}, and \texttt{lp85} employ click-based control, enabling spatial interaction by allowing the agent to click any pixel location in the frame, yielding an action space of size $|\mathcal{A}| = 64 \times 64 = 4{,}096$. Private games (\texttt{sp80} and \texttt{as66}) introduce combined control schemes that integrate both arrow and click inputs, resulting in action spaces of size $|\mathcal{A}| = 4{,}100$.

The dramatic difference in action space cardinality between control schemes poses a fundamental challenge: click-based games present over 1,000 times more possible actions at each state than arrow-based games, making exhaustive exploration intractable without intelligent action selection.




\subsection{Task Structure and Mechanics}

Each game in ARC-AGI-3 embodies a distinct set of mechanics and objectives that agents must discover through interaction. The only feedback signal is level completion: the environment advances to the next level when the agent satisfies (unknown) winning conditions, or resets to the beginning when the step limit expires.

Within each game, levels progressively add new elements while retaining earlier ones. For example, level 1 of \texttt{ls20} requires basic movement and the use of the transformer object to activate the exit door by adjusting the shape of a key, level 2 adds energy palletes to refill the number of steps remaining, level 3 introduces color dimension to the key, and so forth, up to level 8, when the agent must manage with only partial observations. This progressive structure mirrors how humans naturally acquire skills in games, but poses challenges for algorithms: knowledge transfer between levels could accelerate learning, but the levels are connected on a highly abstract level.

The released games operate deterministically: the same action taken from the same state always produces the same outcome. This property enables systematic state-space exploration strategies and graph-based representations of explored states. However, determinism does not imply simplicity; the complexity arises from the large state and action spaces and the lack of prior knowledge about which actions lead toward goal states.

\section{Methods} 
Our approach comprises two primary components: a Frame Processor for extracting key visual features and a Level Graph Explorer for systematic state-space exploration.

\subsection{Frame Processor} The Frame Processor reduces irrelevant visual variability and directs exploration toward actionable regions of the game environment through the following operations:

\subsubsection{Image Segmentation.} Each frame is segmented into single-color connected components, establishing the foundation for identifying distinct visual elements that may constitute interactive objects.

\subsubsection{Status Bar Detection and Masking.} To prevent conflation of environment features with user interface components, the processor identifies and masks probable status bars. This preprocessing substantially reduces the number of recognized states.

\subsubsection{Priority-Based Action Grouping.} For click-controlled games, visual segments are stratified into five priority tiers based on their likelihood of representing interactive buttons or objects. Prioritization is determined by segment size, morphological features, and color salience. The lowest priority tier encompasses segments identified as probable status bars, ensuring their exploration only after exhausting higher-priority alternatives.

\subsubsection{State Hashing.} The processor generates a hash representation of the masked image, serving as a unique identifier for the current game state. This hash facilitates efficient state tracking and duplicate detection during graph exploration.

\subsection{Level Graph Explorer}
The Level Graph Explorer maintains a directed graph representation of the explored state space, where nodes correspond to unique game states and edges encode action-induced state transitions.

\subsubsection{Graph Structure.}
For each discovered state (graph node), the explorer maintains:
\begin{itemize}
    \item The action space $\mathcal{A}$ identifiers of connected components for spatial interaction games such as \texttt{ft09/cv33}, keyboard inputs for games such as \texttt{ls20})
    \item For each action $a \in \mathcal{A}$: priority level $\pi(a)$, exploration status, transition outcome, successor state, and minimal distance to the nearest unexplored frontier
\end{itemize}

\subsubsection{Action Selection Strategy.}
The explorer implements a hierarchical action selection policy that progressively expands the search space, as shown in Algorithm \ref{alg:act_sel}.

\begin{algorithm}
\caption{Hierarchical Action Selection}
\label{alg:act_sel}
\begin{algorithmic}
\REQUIRE Current state $s$, priority threshold $p$
\IF{$\exists$ untested actions with priority $\pi(a) \leq p$ in state $s$}
    \STATE Select uniformly at random an untested action $a$ where $\pi(a) \leq p$ from $s$
    \STATE Execute action and update graph with observed transition
\ELSIF{$\exists$ reachable state $s'$ with untested actions where $\pi(a) \leq p$}
    \STATE Select action minimizing distance to reachable state $s'$ with untested actions at priority $\leq p$
    \STATE Execute selected action
\ELSE
    \STATE Increment priority threshold: $p \leftarrow p + 1$
    \STATE Recurse from current state $s$ with updated priority $p$
\ENDIF
\end{algorithmic}
\end{algorithm}

This policy ensures systematic exploration of high-salience actions prior to considering lower-priority alternatives, thereby focusing computational resources on likely-relevant state-action pairs.

\subsubsection{Frontier Management.}
The explorer maintains the shortest-path distances from each explored state to frontier states, those containing untested actions. These distance metrics always guide traversal toward unexplored regions.

\section{Baselines}

We evaluate our approach against two baseline methods to demonstrate the effectiveness of structured exploration.

\subsubsection{Random Agent.} A simple baseline that selects actions uniformly at random from the available action space at each step. This baseline provides a lower bound on performance and demonstrates the difficulty of solving tasks through undirected exploration alone.

\subsubsection{LLM+DSL.} We compare against the best-performing LLM-based solution on the leaderboard~\cite{fluxon_arc_2025}, which combines GPT-4.1 with domain-specific language (DSL) programming. The approach observes game frames and generates Python code to interact with the environment, attempting to discover game mechanics through programmatic reasoning. Despite using a frontier LLM, this approach demonstrates the current limitations of LLM-based methods for interactive reasoning tasks.

Because each environment step is gated by an LLM call, it is severely interaction-limited: within the evaluation budget, it produces only about 4,000 interactions per game, compared to the 96,000 steps that are in principle allowed. To avoid high LLM usage costs, we do not re-run this baseline; instead, we report the results from its official evaluation on the private games, with the limitation that only a single aggregate score is available and no results are reported on the public games.

\section{Results}

\begin{figure}[t]
  \centering
  \includegraphics[width=0.48\textwidth]{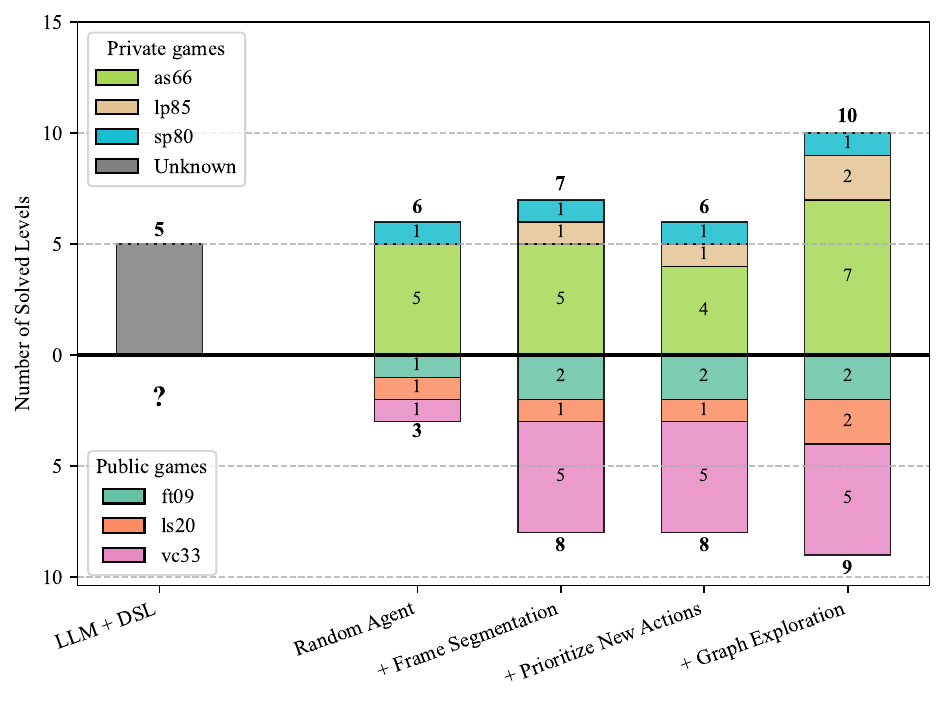}
  \caption{Effect of progressively adding method components to a random agent, compared with the LLM+DSL baseline. For each configuration, the stacked bar above the horizontal axis shows the total number of solved levels across the three private games, and the stacked bar below shows the total across the three public games. Colors indicate how many levels are solved in each individual game. The rightmost bars correspond to the full method. All non-LLM configurations report the median over 5 runs, whereas the LLM+DSL baseline is shown as a single result taken from the official challenge evaluation.}
  \label{fig:results}
\end{figure}

We evaluated our graph-based exploration method on all six ARC-AGI-3 games. Figure~\ref{fig:results} reports an incremental component-addition analysis: starting from a random agent, we cumulatively add components and measure the total levels solved across games; the LLM+DSL baseline is included for comparison. Here, to ensure a fair comparison with the LLM-based baseline, all methods are capped at 4,000 interactions per game. All non-LLM configurations report the median over 5 runs, whereas the LLM+DSL baseline is shown as a single result taken from the official challenge evaluation.

The random agent and LLM+DSL baseline solve 6 and 5 levels on the private games, respectively, meaning that the LLM-based method underperforms even a random policy. The random agent also solves 3 levels across the public games.

Adding frame segmentation to random exploration slightly increases performance on the private games, making it possible to solve one level of \texttt{lp85}. It also significantly improves performance on the public games, solving 5 levels on \texttt{vc33} and 2 levels on \texttt{ft09}.

When, in each state, untested actions are favored without full state-graph exploration, performance slightly decreases on as66, and the method is able to solve only 4 levels.

Our complete approach solves 19 levels with an interaction limit of 4,000: 2 on \texttt{ft09}, 2 on \texttt{ls20}, 5 on \texttt{vc33}, 1 on \texttt{sp80}, 2 on \texttt{lp85}, and 7 on \texttt{as66}.

In a full 8-hour run, across 5 independent runs, our method solves a median of 16 levels on the private games and 14 levels on the public games (see Figure~\ref{fig:levels_by_steps} in the appendix). Per-level performance is reported in Tables~\ref{tab:results_public} and~\ref{tab:results_private} in the appendix.

On the official ARC-AGI-3 challenge evaluation, the submitted model solves 12 levels on the private games while still ranking 3rd by the number of solved levels. This discrepancy is due to an implementation bug in how reset-inducing actions are handled (see Discussion).


\section{Discussion}

Our graph-based exploration method demonstrates that structured state-space navigation with visual prioritization significantly outperforms both random exploration and frontier LLMs with access to code writing and execution on ARC-AGI-3. 

\subsubsection{Performance Analysis.} The method excelled on games where visual salience aligned with interactive elements (\texttt{vc33}, \texttt{as66}). Performance degraded on games with extremely large state spaces (\texttt{ft09} levels 6+, \texttt{ls20} levels 3+), where exhaustive exploration becomes computationally intractable. The improvement over LLM+DSL baselines suggests that structured exploration provides a more reliable foundation for interactive reasoning than pure language-model-based approaches, which struggle to form and test hypotheses systematically.

The discrepancy between the official ARC-AGI-3 evaluation and our re-runs is due to an implementation bug in the handling of reset events. Actions that triggered a reset were not marked as tested in the game graph. Consequently, when such a state–action pair was the nearest remaining untested edge in the graph from the starting node, the agent repeatedly selected it, resetting the game and effectively entering a loop.

\subsubsection{Limitations.} The method faces two fundamental constraints. First, computational requirements grow linearly with state space size, limiting scalability to levels with moderate complexity. Second, the approach assumes deterministic, fully observable environments and would fail under stochasticity or partial observability.

\subsubsection{Future Directions.} While the first-place solution on the leaderboard~\cite{smit_driessmitarc3-solution_2025} achieved superior performance with a learning-based approach, it did not incorporate structured exploration strategies. A natural next step is to integrate our graph-based exploration framework with adaptive learning algorithms.
Such hybrid approaches could leverage graph representations to guide model training and action selection, while learned world models or policies could improve sample efficiency through generalization. The key challenge remains the sparse reward signal and limited training data, making it essential to develop methods that can effectively transfer knowledge across levels while maintaining systematic exploration coverage.
\bibliography{arcagi3}

\clearpage 
\appendix 

\begin{figure*}[t]
  \section{Appendix A: ARC-AGI-3 Games}
  \centering
  \setlength{\tabcolsep}{6pt}
  \begin{tabular}{ccc}
    \includegraphics[width=0.28\textwidth]{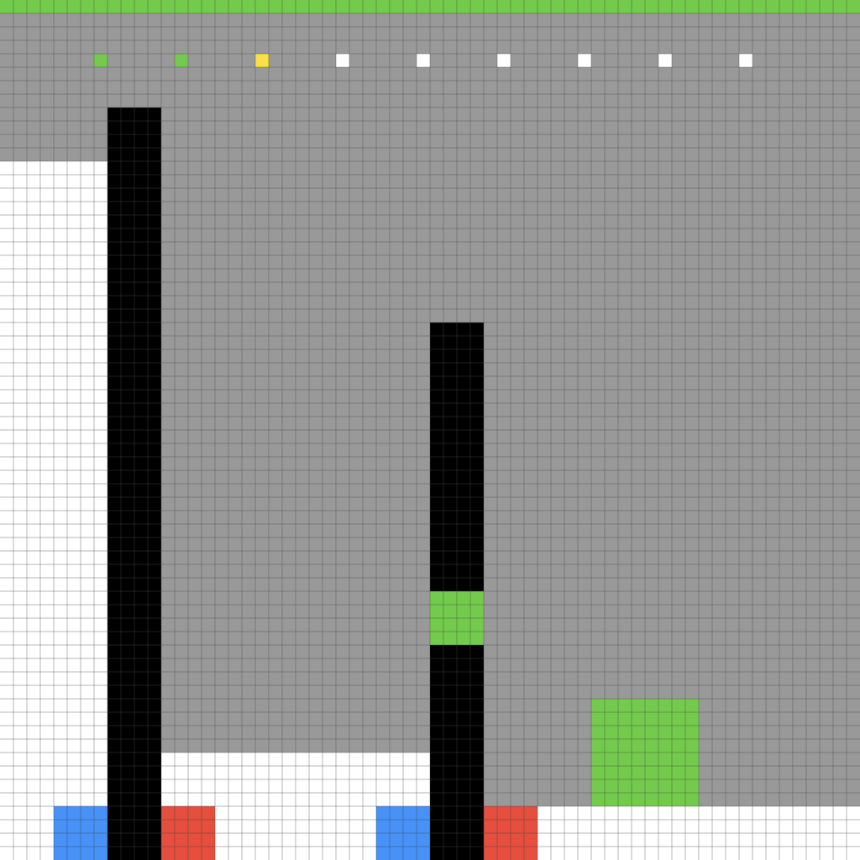} &
    \includegraphics[width=0.28\textwidth]{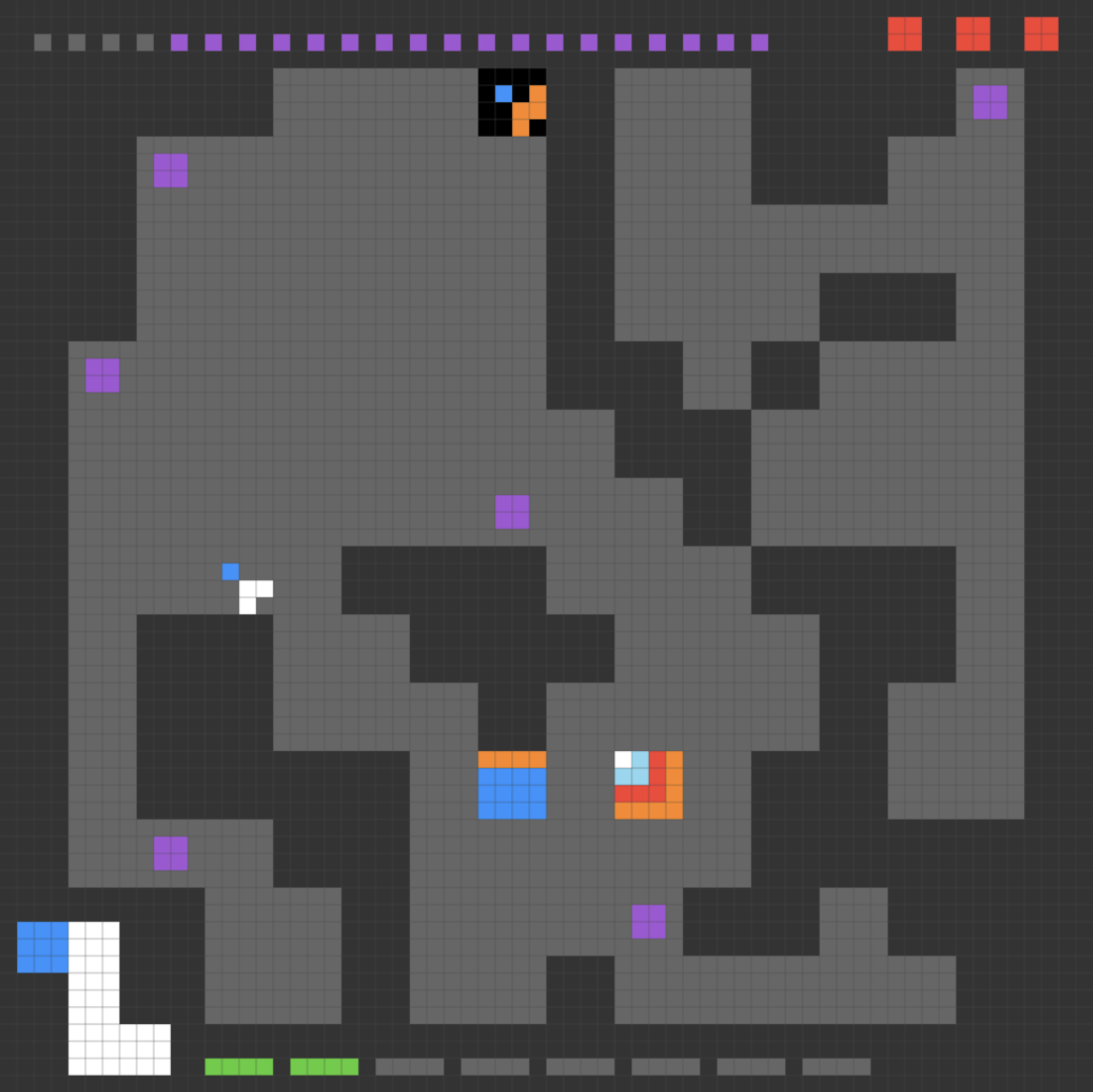} &
    \includegraphics[width=0.28\textwidth]{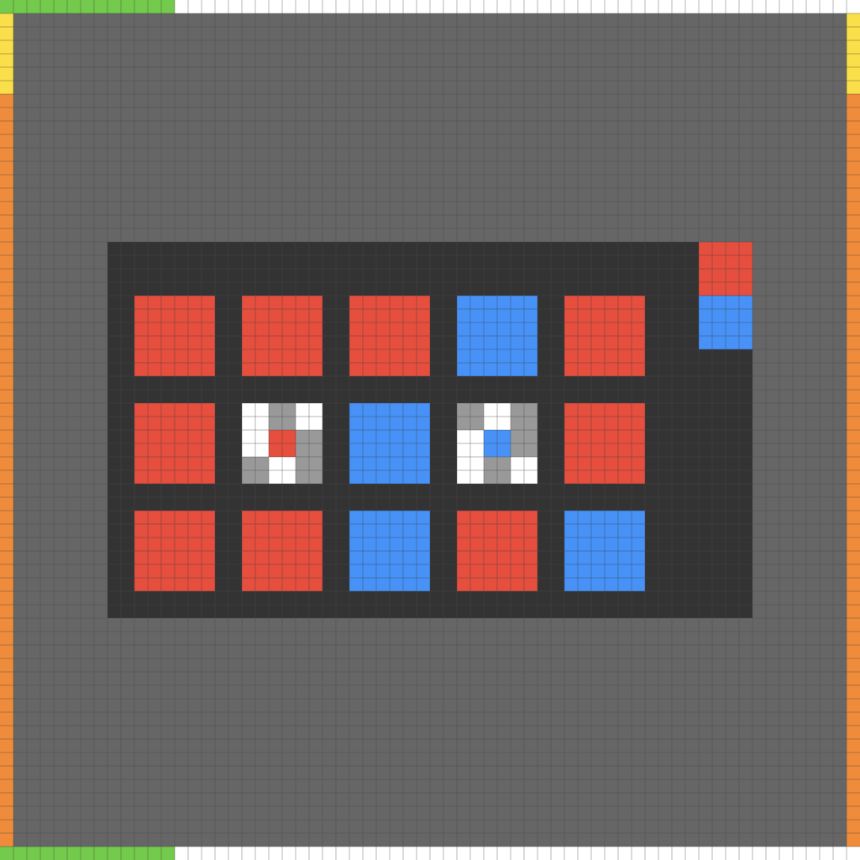} \\
    \includegraphics[width=0.28\textwidth]{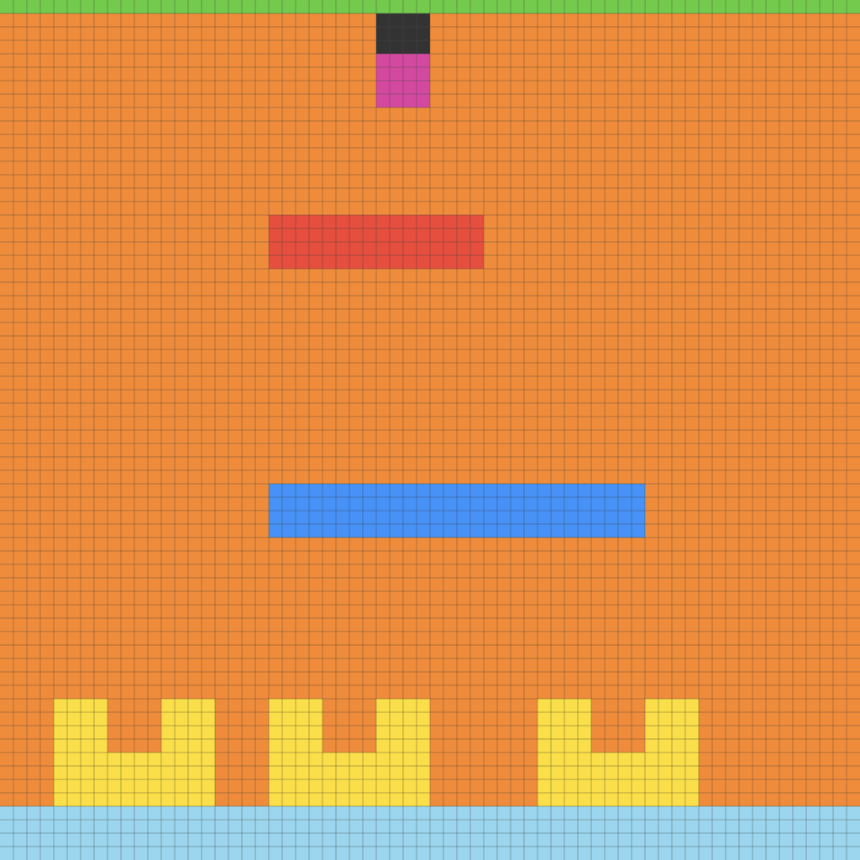} &
    \includegraphics[width=0.28\textwidth]{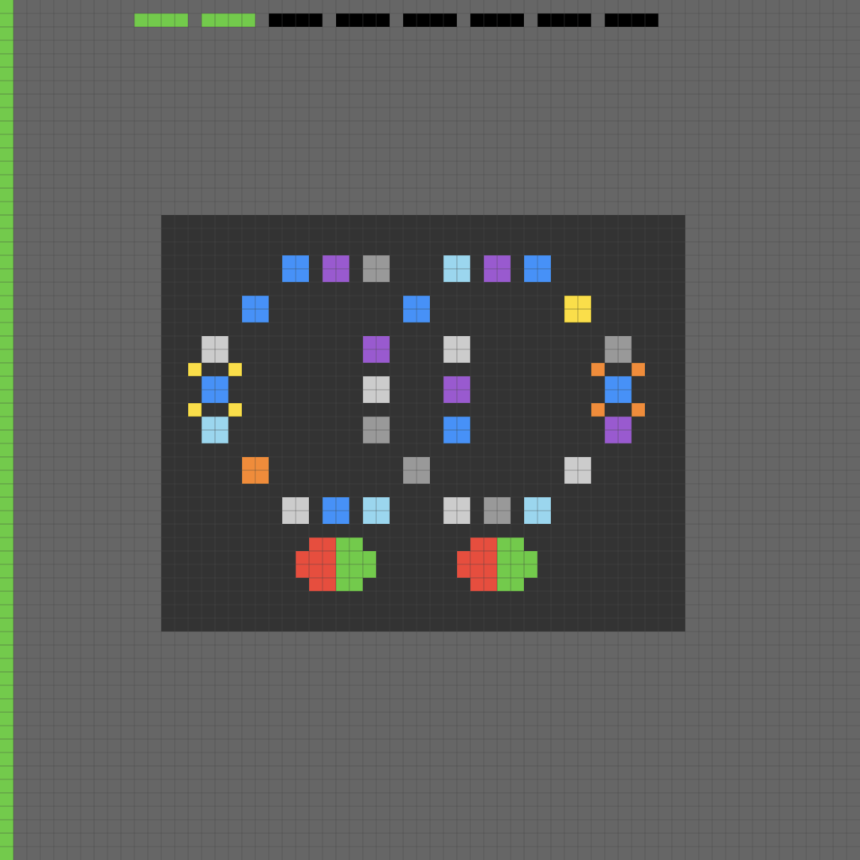} &
    \includegraphics[width=0.28\textwidth]{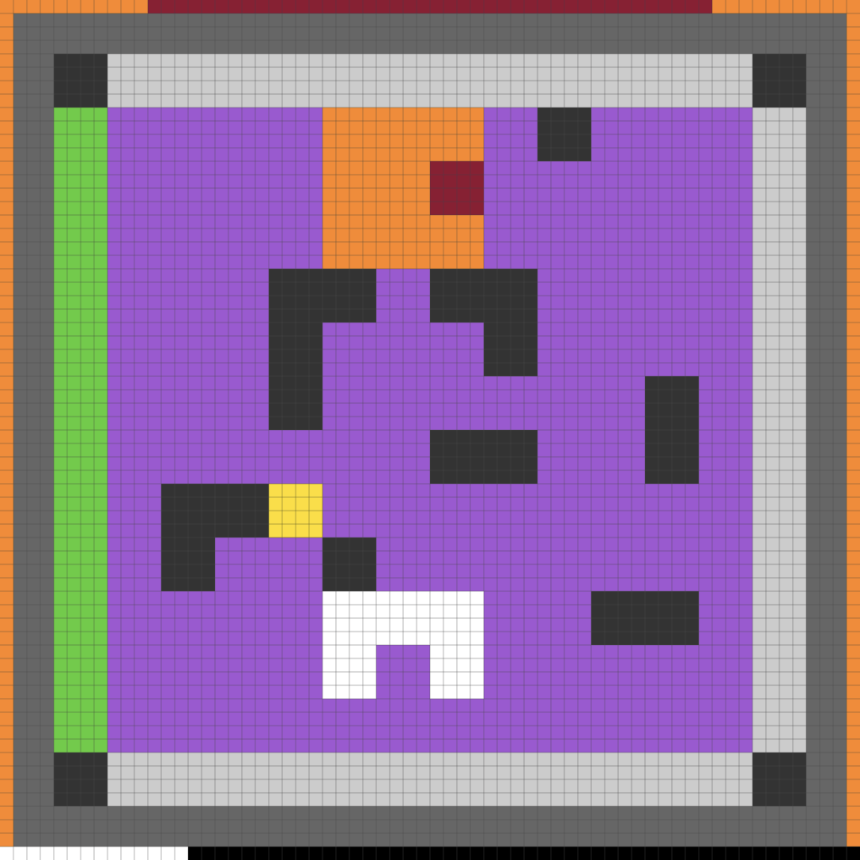} \\
  \end{tabular}
  \caption{Top row (Public set): \texttt{vc33}, \texttt{ls20}, \texttt{ft09}. Bottom row (Private set): \texttt{sp80}, \texttt{lp85}, \texttt{as66}.}
  \label{fig:games}
\end{figure*}

\clearpage

\begin{figure*}[t]
  \section{Appendix B: Per-Level Performance Statistics}
  \centering

    \includegraphics[width=0.9\textwidth]{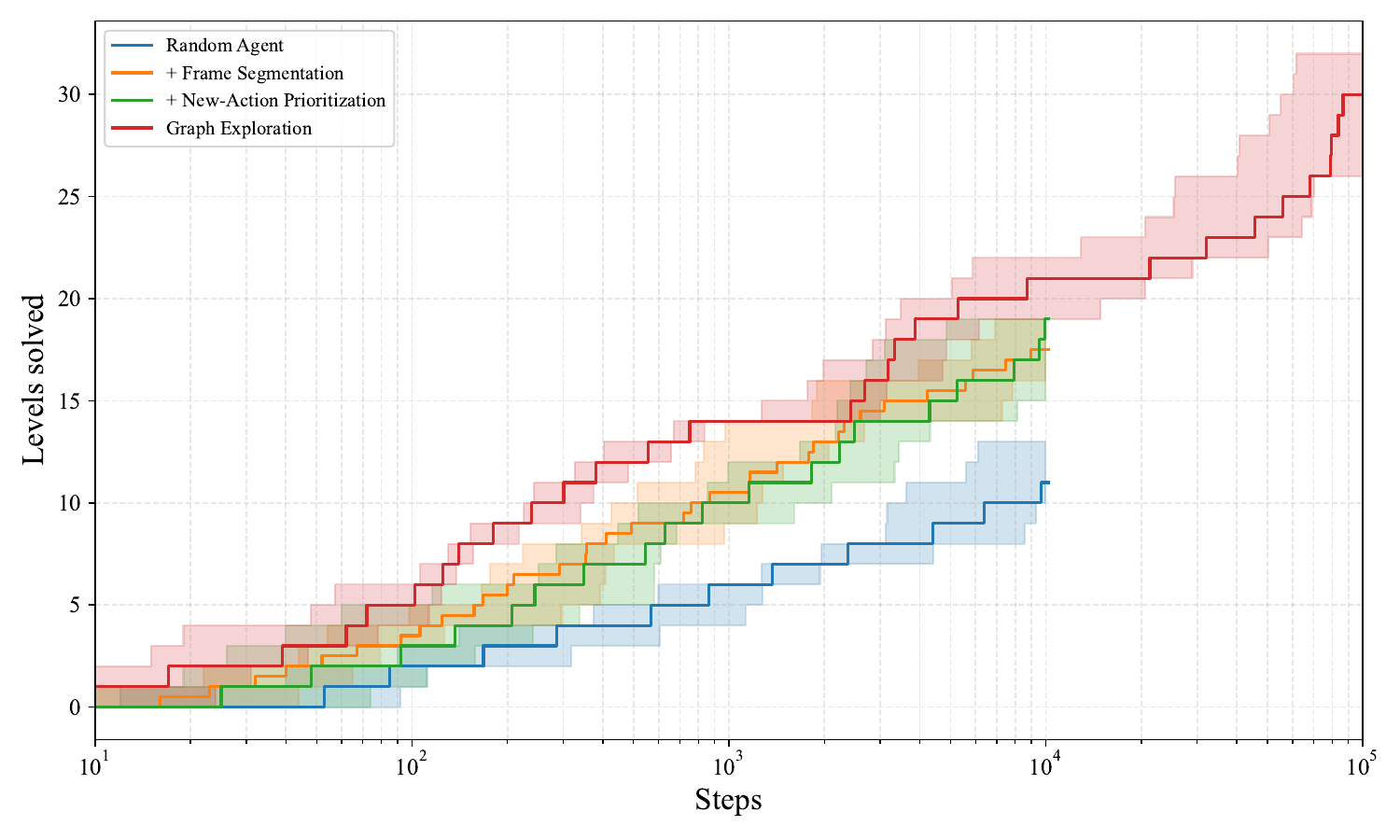}
  \caption{Levels solved as a function of environment steps for four methods: Random Agent, Random + Frame Segmentation, Random + Segmentation + New-Action Prioritization, and the full Graph Exploration method. The x-axis is logarithmic; each line shows the median over 5 runs and the shaded region shows the minimum–maximum range. Intermediate variants are shown up to 10,000 environment steps, while the Graph Explorer is plotted over the full evaluation budget.}
  \label{fig:levels_by_steps}
\end{figure*}

\begin{table*}[h]
\centering
\caption{Per-level results on public games (\texttt{ft09, ls20, vc33}). For each game and level, we report the number of steps to solve the level (\textbf{Stp}), summarized as the median together with the minimum and maximum over 5 runs, and the solve rate (\textbf{SR}) over the same 5 runs. We use `NS` when a level is never solved within the step budget, and `-` when there is no such level for a given game.}
\label{tab:results_public}
\begin{tabular}{|c|cc|cc|cc|}
\hline
 & \multicolumn{2}{c|}{\textbf{ft09}} & \multicolumn{2}{c|}{\textbf{ls20}} & \multicolumn{2}{c|}{\textbf{vc33}} \\
\hline
\textbf{Level} & \textbf{Stp} & \textbf{SR} & \textbf{Stp} & \textbf{SR} & \textbf{Stp} & \textbf{SR} \\
\hline
1  & 125  & 1   & 124  & 1   & 9        & 1 \\
   & $[48;340]$      &     & $[72;140]$       &     & $[5;24]$         &   \\
2  & 177  & 1   & $3.2 \times 10^{3}$ & 1   & 7        & 1 \\
   & $[5;433]$       &     & $[1.9 \times 10^{3};4.9 \times 10^{3}]$ & & $[4;19]$         &   \\
3  & $2.0 \times 10^{4}$ & 1 & NS    & 0   & 36       & 1 \\
   & $[3.0 \times 10^{3};2.5 \times 10^{4}]$ & &     &     & $[9;96]$         &   \\
4  & NS    & 0   & NS    & 0   & 321      & 1 \\
   &               &     &           &     & $[298;541]$    &   \\
5  & NS    & 0   & NS    & 0   & 287      & 1 \\
   &               &     &           &     & $[260;349]$    &   \\
6  & NS    & 0   & NS    & 0   & $6.9 \times 10^{4}$ & 0.8 \\
   &               &     &           &     & $[5.4 \times 10^{4};8.3 \times 10^{4}]$ & \\
7  & NS    & 0   & NS    & 0   & $4.7 \times 10^{3}$ & 0.8 \\
   &               &     &           &     & $[1.5 \times 10^{3};5.5 \times 10^{3}]$ & \\
8  & NS    & 0   & NS    & 0   & 917      & 0.8 \\
   &               &     &           &     & $[627;929]$    &   \\
9  & NS    & 0   & -    & -   & NS        & 0 \\
   &               &     &           &     &               &   \\
10 & NS    & 0   & -    & -   & -        & - \\
   &              &     &           &     &               &   \\
\hline
\end{tabular}
\end{table*}

\begin{table*}[h]
\centering
\caption{Per-level results on private games (\texttt{sp80, lp85, as66}). Conventions as in Table~\ref{tab:results_public}.}
\label{tab:results_private}
\begin{tabular}{|c|cc|cc|cc|}
\hline
 & \multicolumn{2}{c|}{\textbf{sp80}} & \multicolumn{2}{c|}{\textbf{lp85}} & \multicolumn{2}{c|}{\textbf{as66}} \\
\hline
\textbf{Level} & \textbf{Stp} & \textbf{SR} & \textbf{Stp} & \textbf{SR} & \textbf{Stp} & \textbf{SR} \\
\hline
1  & 227                 & 1   & 143                 & 1   & 39                  & 1 \\
   & $[153;373]$         &     & $[106;181]$         &     & $[13;47]$           &   \\
2  & $3.6 \times 10^{4}$ & 1   & $2.9 \times 10^{3}$ & 1   & 44                  & 1 \\
   & $[2.5 \times 10^{4};5.0 \times 10^{4}]$ & & $[1.1 \times 10^{3};3.2 \times 10^{4}]$ & & $[24;65]$ &   \\
3  & $3.9 \times 10^{4}$ & 0.4 & $1.7 \times 10^{4}$ & 1   & 123                 & 1 \\
   & $[3.6 \times 10^{4};4.2 \times 10^{4}]$ & & $[1.0 \times 10^{4};8.2 \times 10^{4}]$ & & $[25;339]$ &   \\
4  & NS                   & 0   & $1.6 \times 10^{3}$ & 1   & 99                  & 1 \\
   &                    &     & $[727;2.0 \times 10^{4}]$ & & $[69;350]$ &   \\
5  & NS                   &    & $4.6 \times 10^{3}$ & 0.8 & $2.2 \times 10^{3}$ & 1 \\
   &                    &     & $[2.2 \times 10^{3};1.4 \times 10^{4}]$ & & $[1.2 \times 10^{3};2.9 \times 10^{3}]$ & \\
6  & NS                   & 0   & $1.3 \times 10^{4}$ & 0.4 & $1.3 \times 10^{3}$ & 1 \\
   &                    &     & $[1.1 \times 10^{4};1.5 \times 10^{4}]$ & & $[112;1.6 \times 10^{3}]$ & \\
7  & NS                   & 0   & 334.5               & 0.4 & 363                 & 1 \\
   &                    &     & $[104;565]$         &     & $[128;670]$         & \\
8  & NS                   & 0   & $9.9 \times 10^{3}$ & 0.2 & $1.3 \times 10^{3}$ & 1 \\
   &                    &     & $[9.9 \times 10^{3};9.9 \times 10^{3}]$ & & $[168;2.9 \times 10^{3}]$ & \\
9  & -                   & -   & -                   & -   & $3.4 \times 10^{3}$ & 1 \\
   &                    &     &                    &     & $[361;8.7 \times 10^{3}]$ & \\
10 & -                   & -   & -                   & -   & -                   & - \\
   &                    &     &                    &     &                    & \\
\hline
\end{tabular}
\end{table*}

\end{document}